\documentclass[10pt, conference]{IEEEtran}
\IEEEoverridecommandlockouts
\IEEEpubid{\makebox[\columnwidth]{979-8-3315-8654-6/25/\$31.00~\copyright2025 IEEE \hfill}
\hspace{\columnsep}\makebox[\columnwidth]{ }}

\IEEEpubidadjcol

\usepackage{amsmath,amssymb,amsfonts}
\usepackage{algorithmic}
\usepackage{subcaption}
\usepackage[numbers]{natbib}
\usepackage{array, longtable, tabularx}
\usepackage{pdflscape}
\usepackage{float}
\usepackage{graphicx}
\usepackage{booktabs}
\usepackage{textcomp}
\usepackage{xcolor}
\def\BibTeX{{\rm B\kern-.05em{\sc i\kern-.025em b}\kern-.08em
    T\kern-.1667em\lower.7ex\hbox{E}\kern-.125emX}}

\usepackage[english]{babel}
\emergencystretch=\maxdimen
\begin{document}

\title{Enhanced Graph Convolutional Network with Chebyshev Spectral Graph and Graph
Attention for Autism Spectrum Disorder Classification}

\DeclareRobustCommand*{\IEEEauthorrefmark}[1]{%
  \raisebox{0pt}[0pt][0pt]{\textsuperscript{\footnotesize\ensuremath{#1}}}}

\author{\IEEEauthorblockN{Adnan Ferdous Ashrafi\IEEEauthorrefmark{1}, Md. Hasanul Kabir\IEEEauthorrefmark{2}}\\
\IEEEauthorblockA{\textit{\IEEEauthorrefmark{1,2}Department of Computer Science and Engineering} \\ \textit{Islamic University of Technology, KB Road, Gazipur -1704, Bangladesh}\\
Email: nazib91@iut-dhaka.edu\IEEEauthorrefmark{1}, hasanul@iut-dhaka.edu\IEEEauthorrefmark{2}}
}

\maketitle

\begin{abstract}
ASD is a complicated neurodevelopmental disorder marked by variation in symptom presentation and neurological underpinnings, making early and objective diagnosis extremely problematic.  This paper presents a Graph Convolutional Network (GCN) model, incorporating Chebyshev Spectral Graph Convolution and Graph Attention Networks (GAT), to increase the classification accuracy of ASD utilizing multimodal neuroimaging and phenotypic data.  Leveraging the ABIDE I dataset, which contains resting-state functional MRI (rs-fMRI), structural MRI (sMRI), and phenotypic variables from 870 patients, the model leverages a multi-branch architecture that processes each modality individually before merging them via concatenation.  Graph structure is encoded using site-based similarity to generate a population graph, which helps in understanding relationship connections across individuals.  Chebyshev polynomial filters provide localized spectral learning with lower computational complexity, whereas GAT layers increase node representations by attention-weighted aggregation of surrounding information.  The proposed model is trained using stratified five-fold cross-validation with a total input dimension of 5,206 features per individual.  Extensive trials demonstrate the enhanced model's superiority, achieving a test accuracy of 74.82\% and an AUC of 0.82 on the entire dataset, surpassing multiple state-of-the-art baselines, including conventional GCNs, autoencoder-based deep neural networks, and multimodal CNNs.  
\end{abstract}


\section{Introduction}

Autism Spectrum Disorder (ASD) is recognized as a complex neurodevelopmental disorder characterized by persistent deficits in social interaction, communication, and restricted, repetitive behaviors and interests. Currently, the diagnosis of ASD is largely based on behavioral observations and clinical interviews, which are performed by clinical experts in neurodevelopmental disorders. However, these methods are considered subjective, and there is an absence of laboratory-based diagnostic tests, making identifying and diagnosing ASD a challenging task. Early detection of ASD and specialized support are crucial for prompt diagnosis and access to essential services and interventions, which can improve the quality of life.

Observation and interview methodologies are two prevalent manual strategies for the identification and diagnosis of ASD.  The Childhood Autism Rating Scale (CARS)~\cite{schopler1980toward} comprises 15 items designed to evaluate ASD.   CARS provides a spectrum of scores to denote ASD levels; for instance, a score of 30-37 indicates moderate ASD, but a score of 38-60 signifies severe ASD.   Conversely, interview-based detection and diagnostic systems~\cite{skuse2004developmental, lord1994autism, johansson2001autistic, lecavalier2005evaluation} rely on discussions with parents or caregivers.  Nonetheless, the manual techniques rely on behavioral symptoms and the observations of parents or caregivers, necessitating the expertise of a physician for accurate assessments.   Consequently, it is unable to collect data on authentic conditions of routine everyday operations.   Furthermore, these methodologies are expensive and labor-intensive~\cite{galliver2017cost, rutter2003autism}. In recent years, non-invasive brain imaging techniques, such as magnetic resonance imaging (MRI), have been explored to identify structural and functional differences between individuals with ASD and typical controls~\cite{stigler2011structural}, aiding in objective diagnosis and identifying neural correlates.

However, utilizing MRI data for ASD detection faces several challenges~\cite{hugdahl2012autism}. Issues include variations in brain connectivity patterns, small sample sizes, and data heterogeneity, particularly from multi-site collections like the Autism Brain Imaging Data Exchange (ABIDE)~\cite{DiMartino2013} dataset, which introduces variability due to different imaging procedures and scanners. Conventional image classification methods have often failed to accurately identify the disorder~\cite{yi2014individuals}, frequently focusing solely on imaging features and overlooking important non-imaging data. Furthermore, traditional techniques often rely on pairwise comparisons and may not fully capture complex interactions or individual characteristics. The inherent biological heterogeneity of autism also poses a significant limitation to classification performance~\cite{an2016genetic}. These challenges underscore the need for improved analysis techniques. 

This study contributes a novel multi-branch Graph Convolutional Network (GCN) architecture that combines Chebyshev Spectral Graph Convolution and Graph Attention Networks (GAT) to classify ASD using multimodal neuroimaging and phenotypic data. By constructing a site-based population graph and processing each modality through specialised branches, the model effectively captures complex inter-subject relationships. Extensive evaluation on the ABIDE I dataset demonstrates state-of-the-art performance, achieving 74.81\% accuracy and 0.82 AUC, with ablation studies confirming the value of graph attention and hyperparameter tuning.

\section{Related Works}
Recent advancements have seen deep learning models applied extensively in the medical imaging field, aiming for improved accuracy compared to classical machine learning methods. These models have achieved performance levels comparable to human experts in various domains, including natural language processing and computer vision.

\subsection{Autoencoders}
One of the first successful application of deep learning in ASD detection utilized deep neural networks combining stacked denoising autoencoders~\cite{Heinsfeld2018} and a fully connected network (FCN) on fMRI data from the ABIDE dataset, achieving a 70\% classification accuracy. Later, an autoencoder followed by a fully connected network (AE-FCN)~\cite{rakic2020improving} along with Ensembles of Multiple Models and Architectures (EMMA), incorporating both functional MRI (fMRI) and structural MRI (sMRI) data to reach 85\% accuracy.

\subsection{GCN}
A graph convolutional neural network (GCN)~\cite{parisot2018disease} that integrates phenotypic information, obtained a best accuracy of 70.4\% on ABIDE fMRI data. Building on the GCN, an edge-variational graph convolutional neural network (EV-GCN)~\cite{huang2020edge} was proposed, which reportedly achieved 81\% classification accuracy. Furthermore, GCN and Ensemble GCN~\cite{Dong2025} on the ABIDE I dataset, obtained 68.6\% and 71.3\% accuracy, respectively. Later, multi-modal GCN (MMGCN)~\cite{Song2024} was introduced, deploying multimodal data from the ABIDE I dataset and obtained an accuracy of 72.37\%. A weight-learning deep GCN~\cite{Wang2023} proposed that deployed pairwise associations of non-imaging data into the already existing DeepGCN classification model. The authors achieved an accuracy of 77.27\% on a partial testing dataset. The Fuzzy MSE-GCN~\cite{Rajaprakash2025} framework was introduced later for ASD detection that integrates imaging and non-imaging phenotypic data into a population graph. This approach achieved an accuracy of 87\% on the ABIDE dataset.

\subsection{Domain Adaptation}

Local Global Multimodal Domain Adaptation (LGMDA) paired with Sparse Adaptive Prior Coupled Dictionary Learning (SACDL) framework~\cite{https://doi.org/10.1002/ima.23110} employs LGMDA to perform multi-source domain adaptation and SACDL for sparse-adaptive coupled dictionary learning to fuse sMRI (regional morphology) and fMRI (functional connectivity) features. The approach enhances ASD classification performance on the ABIDE dataset, achieving an accuracy of 75.32\%. 

\subsection{Convolutional Neural Network Based Systems}

DeepMNF~\cite{Abbas2023} a multi-modal CNN on ABIDE I dataset was deployed by Abbas et al., and obtained an accuracy of 75.14\% on a partial testing set. The authors utilized both sMRI and fMRI data available from the dataset. Saponaro et al. utilized a multimodal deep learning model employing a joint fusion approach~\cite{Saponaro2024} was developed by integrating harmonized structural and functional MRI data. This approach achieved an accuracy of 85\% $\pm$ 0.12 and AUC of 0.78 $\pm$ 0.04 on a partial testing dataset.

\subsection{FCN and ViT based Systems}
TinyViT~\cite{gupta2025cross} deployed by Gupta et. al, exploited the generalizability of transfer learning using a student-teacher network on the ABIDE I dataset. The authors achieved an accuracy of 76.62\% on a partial testing dataset. Du et al.~\cite{10821533} deployed a fully connected network (FCN) based system as the preliminary classifier to achieve an accuracy of 68.98\% on the testing dataset.

\section{Materials and Methods}

For this study, a multi-modal Graph Convolutional Neural (GCN) network architecture has been adopted for classification tasks, specifically tailored to integrate different types of brain imaging and phenotypic data. The architecture employs graph convolutional layers alongside an attention mechanism to leverage both feature information and graph structural relationships.

\subsection{Dataset}
The ABIDE I~\cite{DiMartino2013} dataset is a significant, open-access neuroimaging collection aimed at enhancing the comprehension of ASD by extensive brain imaging analysis.  It contains data from over 1,100 individuals, including both ASD cases and typically developing (TD) controls, gathered across 20+ international sites.  The dataset comprises resting-state functional MRI (rs-fMRI) and structural MRI (sMRI) images, plus phenotypic information such as age, sex, IQ, and diagnostic scores (e.g., ADOS, ADI-R).  ABIDE provides cross-site harmonization and is frequently used for deep learning-based categorization, brain connection studies, and functional network modeling.  The dataset is separated into ABIDE I and ABIDE II, with ABIDE II enhancing subject variety and improving phenotypic coverage.


\begin{figure*}
    \centering
    \includegraphics[width=\linewidth]{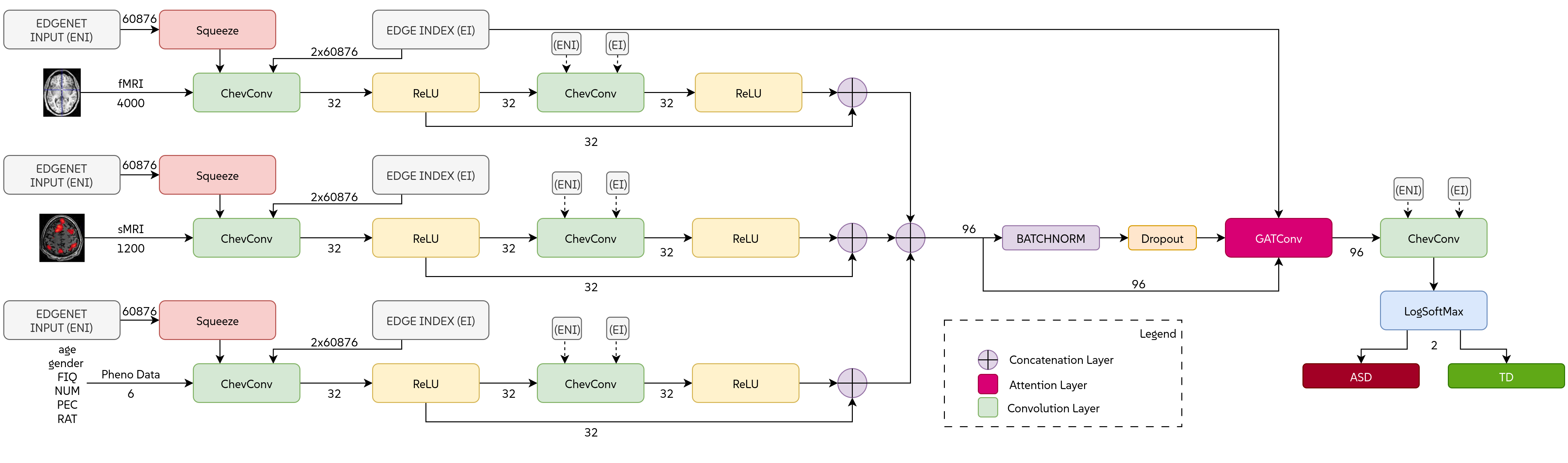}
    \caption{The proposed enhanced GCN architecture with Graph Attention}
    \label{fig:asd_gcn_model}
\end{figure*}
\subsection{Chebyshev Spectral Graph Convolution}

The Chebyshev Spectral Graph Convolution layer~\cite{defferrard2016convolutional}, a fundamental element of the proposed graph Convolutional Neural Network (CNN) framework, makes substantial contributions to the generalization of CNNs to irregular domains such as graphs.  One of its primary contributions is the provision of filters that are rigorously localized. This means that the influence of a filter at a central vertex is limited to a disk of radius K hops, a property that is not inherent in non-parametric spectral filters.  In order to accomplish this localization, the filter is parametrized by a polynomial of order K-1 using the Chebyshev expansion, where K denotes the filter's support size.


\subsection{Graph Attention Networks}

By utilizing masked self-attentional layers, the Graph Attention Network (GAT)~\cite{velivckovic2017graph} presents a revolutionary neural network architecture for graph-structured data, directly addressing the shortcomings of previous approaches that relied on graph convolutions or their approximations.   One of its key characteristics is that it enables nodes to implicitly assign different weights to nodes in their vicinity without the need for costly matrix operations like inversion or previous knowledge of the network architecture.   This is performed by use of a graph attentional layer in which the characteristics of each node $i$ and its neighbors $j \in N_i$ are first subjected to a common linear transformation (parameterized by weight matrix $W$).   The significance of node $j$'s attributes to node $i$ is represented by attention coefficients $e_{ij}$, computed by a common attentional mechanism $a$.   The new feature representation for node $i$ is produced by normalizing these coefficients using a softmax function across the neighborhood of node $i$ to yield $\alpha_{ij}$.  These coefficients are subsequently applied as weights in a linear combination of the transformed neighbor attributes.

It is computed as
\begin{equation}
\mathbf{x}^{\prime}_i = \sum_{j \in \mathcal{N}(i) \cup \{ i \}}
\alpha_{i,j}\mathbf{\Theta}_t\mathbf{x}_{j},    
\end{equation}

\subsection{Enhanced GCN Model Architecture}

Existing GCN architectures consider all modalities as the same and thus constructs a graph edge network in a sequential layering model. However, all of the modalities are different in the ABIDE I dataset, and thus special attention is required for each type of modality in order to extract maximum amount of features from them. The proposed model addresses this issue by utilizing a multi-branch Graph Convolutional Network (GCN) designed for binary classification of ASD and TD individuals. It integrates multimodal brain imaging and phenotypic data using Chebyshev Spectral Graph Convolutions and Graph Attention Networks (GAT) as depicted in Figure \ref{fig:asd_gcn_model} to learn from both relational and feature-based representations. The pipeline comprises several interconnected stages, including data preprocessing, feature extraction, population graph construction, and model training with cross-validation.

\subsubsection{Input Modality}
The model processes three modalities from 870 subjects: functional MRI (fMRI) with dimensions $870 \times 4000$, structural MRI (sMRI) with $870 \times 1200$, and phenotypic data (age, gender, FIQ, NUM, PEC, and RAT) with $870 \times 6$. Each modality is fed into a dedicated branch for specialized graph convolutional transformations.

\subsubsection{Branched Graph Convolution}
Each branch uses graph connectivity defined by an EDGE INDEX and EDGENET INPUT (60,876 pairwise connections). After a dimensional alignment via a Squeeze operation, two successive Chebyshev Graph Convolutional layers (ChebConv) are applied, each followed by ReLU activation. Residual connections between the layers support gradient flow and training stability.

\subsubsection{Branch Concatenation}
The outputs of all branches (each $870 \times 32$) are concatenated into a unified embedding of size $870 \times 96$, followed by Batch Normalization and Dropout to stabilize training and prevent overfitting.

\subsubsection{Graph Attention}
A Graph Attention Convolution (GATConv) layer is applied to the fused representation to dynamically weight the importance of neighboring nodes. Its output ($870 \times 96$) is added to its input via a residual connection for enhanced expressiveness.

A final ChebConv layer reduces the representation to $870 \times 2$, matching the binary classification goal. A LogSoftmax activation converts this into log-probabilities for ASD and TD.

\begin{table*}[htbp]
\centering
\caption{Ablation study of different enhanced GCN modifications}
\begin{tabularx}{\textwidth}{lcc*{2}{>{\centering\arraybackslash}X}*{2}{>{\centering\arraybackslash}X}}
\toprule
\textbf{Model} & \textbf{Test ACC} & \textbf{Test AUC} & \multicolumn{2}{c}{\textbf{Test ACC Increment (\%)}} & \multicolumn{2}{c}{\textbf{Test AUC Increment (\%)}} \\
\cmidrule(lr){4-5} \cmidrule(lr){6-7}
 & & & \textbf{GCN} & \textbf{Ensemble GCN} & \textbf{GCN} & \textbf{Ensemble GCN} \\
\midrule
EGCN w/o GAT & 70.14\% & 0.78 & 1.54\% & -1.16\% & 3.79\% & -0.01\% \\
EGCN w GAT & 71.24\% & 0.80 & 2.64\% & -0.06\% & 5.45\% & 1.65\% \\
EGCN w GAT w/o HPT & 72.43\% & 0.81 & 3.83\% & 1.13\% & 6.33\% & 2.53\% \\
\textbf{EGCN w GAT w HPT} & \textbf{74.82\%} & \textbf{0.82} & 6.21\% & 3.51\% & 7.33\% & 3.53\% \\
\bottomrule
\end{tabularx}
\label{tab:ablation_tabularx}
\end{table*}

\subsubsection{Final Output}
The model outputs a $870 \times 2$ matrix containing log-probabilities for each subject. This architecture effectively captures complex inter-subject and multi-modal relationships, demonstrating the efficiency of graph-based deep learning for population-scale neuroimaging analysis.

\section{Experimental Setup}

The experimental framework is designed to evaluate the proposed enhanced GCN architecture for classifying ASD using neuroimaging and phenotypic data. The study is conducted using data from the ABIDE I dataset, which includes resting-state functional MRI (fMRI), structural MRI (sMRI), and six types of phenotypic information. 

\subsection{Model Training}

Model training employs the Stochastic Gradient Descent (SGD) optimizer with momentum and Nesterov acceleration. The learning rate is regulated using a Cyclic Learning Rate scheduler with triangular mode. The training loop is deployed for 200 epochs per fold. After each epoch, a validation is conducted on the remaining fold data using a cross-validation method. The model epoch yielding the best validation accuracy is saved, and its performance on the test set is logged. This comprehensive experimental setup ensures rigorous validation of the proposed model, leveraging the strengths of graph-based representation learning and multi-modal feature integration to advance the classification of ASD from neuroimaging and behavioral data.

\subsection{Testing}

As the experiment was conducted using a five-fold cross-validation method, the testing phase included each sample from the training set. The testing was evaluated using generic evaluation metrics like accuracy and AUC, to ensure comparability with state-of-the-art works in this domain.

\subsection{Evaluation Metrics}

The model's performance is assessed using numerous measures, including accuracy and the Area Under the Receiver Operating Characteristic Curve (AUC-ROC), each offering complimentary insights on classification quality.  Accuracy as seen in \ref{eq:acc} is defined as the ratio of accurately predicted labels to the total number of predictions, computed as:

 \begin{equation} \text{Accuracy} = \frac{TP + TN}{TP + TN + FP + FN} \label{eq:acc} \end{equation}

 where $TP$, $TN$, $FP$, and $FN$ signify true positives, true negatives, false positives, and false negatives, respectively.  While accuracy gives a basic measure of overall correctness, it may be skewed in unbalanced datasets.  To account for probabilistic performance, the model additionally assesses the Negative Log-Likelihood (NLL) Loss as found in (\ref{eq:loss}), which measures the divergence between predicted log-probabilities and the real class labels.  NLL is given by:

\begin{equation}
    \text{NLL Loss} = -\sum_{i=1}^{N} \log p(y_i \mid x_i)
    \label{eq:loss}
\end{equation}

where $p(y_i \mid x_i)$ represents the predicted probability for the true label $y_i$ of instance $x_i$, and $N$ is the total number of samples. Lastly, the AUC-ROC metric as formulated in \ref{eq:auc}  captures the model's ability to discriminate between classes by plotting the True Positive Rate (TPR) against the False Positive Rate (FPR) across various threshold settings. The AUC is formally expressed as:

\begin{equation}
    \text{AUC} = \int_{0}^{1} \text{TPR}(FPR^{-1}(x)) \, dx
    \label{eq:auc}
\end{equation}

where TPR and FPR are defined as $\text{TPR} = \frac{TP}{TP + FN}$ and $\text{FPR} = \frac{FP}{FP + TN}$, respectively.

\subsection{Hyperparameter settings}

The model training method involves careful specification and, in certain instances, adjustment of multiple important hyperparameters to enhance performance.  The Graph Convolutional Network (GCN) architecture was set with an input dimension of 5206, a hidden dimension of 32, and an output dimension of 2.  For its Chebyshev convolutional layers, the orders $K_1$ and $K_2$ were set to 2 and 5, respectively.  An initial dropout rate of 0.5 was adopted during single-fold training. For the optimization method, the training phase consistently applied the Stochastic Gradient Descent (SGD) optimizer.  This SGD optimizer was configured with a learning rate of $1e^{-3}$, a momentum of 0.8, Nesterov acceleration, and a weight decay of 1.0.  Learning rate changes were controlled by a cyclic learning rate scheduler, which had a base learning rate of $1e^{-3}$, a maximum learning rate of $1e^{-7}$, a triangular mode, and step sizes of 500 for up-cycle and 300 for down-cycle phases.  Additionally, gradient clipping with a maximum norm of 2.0 was applied to stabilize training.

\begin{table*}[htbp]
\centering
\caption{Performance Comparison of Different GCN-based Models}
\label{tab:gcn}
\begin{tabularx}{0.9\textwidth}{lcc*{2}{>{\centering\arraybackslash}X}*{2}{>{\centering\arraybackslash}X}}
\toprule
\textbf{Ref.} & \textbf{Model} & \textbf{Testing Subset} & \textbf{Test ACC} & \textbf{Test AUC}\\
\midrule
Dong et. al.\cite{Dong2025} & GCN & Complete & 68.60\% & 0.74  \\
Dong et. al.\cite{Dong2025} & Ensemble GCN & Complete & 71.30\% & 0.78 \\
Heinsfeld et. al.\cite{Heinsfeld2018} & Autoencoder based DNN & Complete & 65.00\% & - \\
Du et. al. \cite{Du2022} & SVM-RFE & Complete & 72.10\% & - \\
Abbas et. al \cite{Abbas2023} & Multi-modal CNN & Partial & 75.14\% & 0.77 \\
Wang et. al \cite{Wang2023} & Weight Learning DeepGCN& Partial & 77.27\% & 0.82 \\
Zhang et. al. \cite{Zhang2024} & LGMDA-SACDL  & Partial & 75.32\% & 0.75 \\
Song et. al \cite{Song2024} & MMGCN & Complete & 72.37\% & 0.76 \\
Saponaro et. al \cite{Saponaro2024} & FR-NN with C-NN& Partial & 85.02\% & 0.78 \\
Ma et. al \cite{Ma2024} & Multi-atlas Fusion Template with GCN & Partial & 75.70\% & 0.73 \\
Gupta et. al \cite{gupta2025cross} & TinyViT & Partial & 76.62\% & - \\
Du et. al \cite{10821533} & FCN based model & Complete & 68.98\% & - \\
\textbf{Proposed} & \textbf{Enhanced GCN With Graph Attention} & Complete & \textbf{74.81\%} & \textbf{0.82} \\
\bottomrule
\end{tabularx}
\end{table*}

\section{Experimental results and analysis}
The state-of-the-art performance quantities are reported as found in their original manuscripts.

\subsection{Ablation Study}
Here in Table~\ref{tab:ablation_tabularx}, a quantitative analysis of different configurations of the model architecture is available. The baseline EGCN without incorporating the Graph Attention Network (GAT) layer achieves a test accuracy of 70.14\% and an AUC of 0.78, reflecting marginal improvements of 1.54\% in ACC and 3.79\% in AUC over the standard GCN. Notably, adding the GAT module (EGCN w GAT) yields a slight performance gain but not the best result in the experimentation, with test accuracy rising to 71.24\% and AUC to 0.80, suggesting that attention mechanisms alone may not consistently improve model generalization. The full model, EGCN with both GAT and hyperparameter tuning (EGCN w GAT w HPT), achieves the highest performance, with a test accuracy of 74.82\% and AUC of 0.82. This corresponds to a substantial improvement of 6.21\% and 7.33\% in ACC and AUC, respectively, over the baseline GCN, and 3.51\% and 3.53\% over the
Ensemble GCN.

\subsection{Comparative Analysis}

Table~\ref{tab:gcn} presents a comprehensive comparison of the proposed Enhanced GCN with Graph Attention model against several state-of-the-art GCN-based and deep learning models for ASD classification. The proposed model achieves a test accuracy of 74.81\% and an AUC of 0.82 on the complete ABIDE dataset, outperforming several baseline and advanced methods. When compared to Dong et al. \cite{Dong2025}, whose standard GCN and Ensemble GCN models achieved 68.60\% and 71.30\% accuracy, respectively, the proposed model exhibits a notable performance gain, particularly in AUC, where it surpasses both baselines by 8 and 4 percentage points, respectively. Additionally, the model outperforms other GCN variants such as MMGCN (72.37\% accuracy, 0.76 AUC) proposed by Song et al. \cite{Song2024}, and FCN-based architectures like that of Du et al. \cite{10821533} with 68.98\% accuracy. Among multimodal models trained on partial datasets, the proposed method demonstrates comparable or superior performance to architectures such as the Multi-modal CNN (75.14\% accuracy, 0.77 AUC) by Abbas et al. \cite{Abbas2023}, and Weight Learning DeepGCN (77.27\% accuracy, 0.82 AUC) by Wang et al. \cite{Wang2023}. Although the model by Saponaro et al. \cite{Saponaro2024} reports a higher accuracy of 85.02\%, it is important to note that it is evaluated on a partial subset, which may limit its generalizability. The proposed model maintains state-of-the-art performance with balanced accuracy and AUC on the complete dataset, affirming the effectiveness of graph attention integration and multi-modal learning in enhancing ASD classification.

\begin{figure}[h!]
    
    \includegraphics[width=\linewidth]{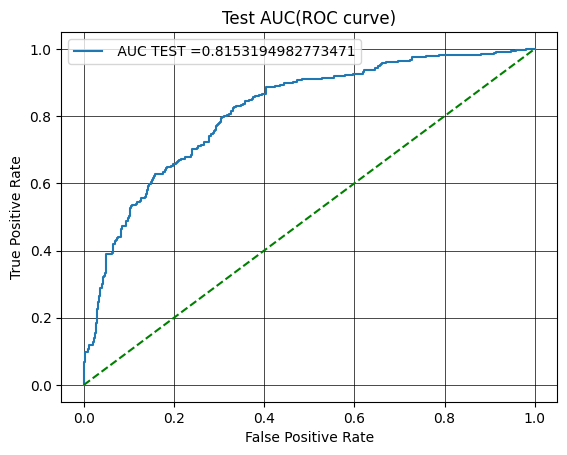}
    \caption{Testing AUC curve}
    \label{fig:test_auc}
    \caption{Qualitative results depicted using the best AUC curve obtained during testing}
    \label{fig:qualitative_result}
\end{figure}

\subsection{Performance Evaluation}

The performance of the presented work is evident from the ROC curves during the experimentation and is presented in Figure~\ref{fig:qualitative_result}. The curves show there was no evident overfitting and the model performed in a quite stable manner to classify between ASD and TD subjects. The final testing AUC is shown in Figure~\ref{fig:test_auc}. An AUC score of 0.82 is evidence that the results are clinically useful and thus have prospective application in real-world settings.


\section{Conclusion}

This paper provides an innovative and successful graph-based deep learning framework for Autism Spectrum Disorder (ASD) classification, utilizing a multi-branch Enhanced Graph Convolutional Network (EGCN) that blends Chebyshev spectral graph convolutions with graph attention mechanisms.

The contributions of this study are diverse. First, the model architecture presents a systematic means of addressing modality-specific feature extraction using parallel ChebConv branches, followed by late fusion and attention-guided refinement. Second, the usage of a population graph based on acquisition site information provides a scalable and generalizable framework for modeling inter-subject interactions, especially in diverse datasets. Third, detailed ablation studies empirically support the separate contributions of the graph attention layer and hyperparameter optimization, showing the relevance of both structural and parametric improvements in deep graph models.

Looking ahead, there are various viable paths to broaden the breadth and effect of this study.  Future studies might examine combining new data modalities, such as genetic or behavioral video data.  Moreover, adaptive or learnable graph-building approaches may provide more dynamic and data-driven connection modeling between individuals. Finally, adding interpretability frameworks into the model may yield neurobiologically significant insights, supporting physicians in identifying crucial biomarkers and processes causing ASD.  Overall, this work offers a solid basis for the continuing improvement of graph-based multi-modal learning in computational neuropsychiatry.

\bibliographystyle{IEEEtran}
\scriptsize
\bibliography{refs}

\end{document}